\documentclass[letterpaper,10pt,journal,twoside]{IEEEtran}
\usepackage{amsmath,amsfonts}
\usepackage{algorithm}
\usepackage{algpseudocode}
\usepackage{array}
\usepackage[caption=false,font=normalsize,labelfont=sf,textfont=sf]{subfig}
\usepackage{textcomp}
\usepackage{stfloats}
\usepackage{url}
\usepackage{verbatim}
\usepackage{graphicx}
\usepackage{cite}
\usepackage{graphics} 
\usepackage{epsfig} 
\usepackage{mathptmx} 
\usepackage{times} 
\usepackage{amsmath} 
\usepackage{amssymb}  

\usepackage{amsthm}
\usepackage[bookmarks=false]{hyperref}
\usepackage{float}
\usepackage{multirow}
\usepackage{booktabs}
\usepackage{paralist, tabularx}
\usepackage[dvipsnames]{xcolor}
\colorlet{blue}{black}
\usepackage{soul}
\usepackage[normalem]{ulem}
\usepackage{balance}
\newtheorem{proposition}{Proposition}

\begin{document}

\title{
Seeing Through Uncertainty: Free-Energy-Inspired Real-Time Adaptation for Robust Visual Navigation
}
\author{Maytus Piriyajitakonkij$^{1,2,3}$, Rishabh Dev Yadav$^{1}$, Mingfei Sun$^{1,\dagger}$, Mengmi Zhang$^{2,3,\dagger}$, and Wei Pan$^{1,\dagger}$
\thanks{Manuscript received: January 30, 2026; Revised: June 15, 2026; Accepted: July 8, 2026.}
\thanks{This paper was recommended for publication by
Editor Tetsuya Ogata upon evaluation of the Associate Editor and Reviewers’ comments. This work is partially supported by the National Research Foundation, Singapore under its NRFF award NRF-NRFF15-2023-0001 and Mengmi Zhang’s Startup Grant from Nanyang Technological University.}
\thanks{$^{1}$Maytus Piriyajitakonkij, Rishabh Dev Yadav, Mingfei Sun, and Wei Pan are with the Department of Computer Science, The University of Manchester, United Kingdom.}%
\thanks{$^{2}$Mengmi Zhang and Maytus Piriyajitakonkij are with the College of Computing and Data Science, Nanyang Technological University, Singapore.}%
\thanks{$^{3}$Mengmi Zhang and Maytus Piriyajitakonkij are also with the Institute of Advanced Intelligence and Computing (IAIC), Agency for Science, Technology and Research (A*STAR), Singapore.}%
\thanks{Project website: \texttt{https://sites.google.com/view/tta-nav}.}
\thanks{$^\dagger$Equal supervision. Corresponding author: \texttt{maytusp@gmail.com}.}%
}

\markboth{IEEE Robotics and Automation Letters. Preprint Version. Accepted July, 2026}%
{Piriyajitakonkij \MakeLowercase{\textit{et al.}}: Free-Energy-Inspired Real-Time Adaptation for Robust Visual Navigation}

\maketitle

\begin{abstract}
Navigation in the natural world is a feat of adaptive inference, where biological organisms maintain goal-directed behaviour despite noisy and incomplete sensory streams. Central to this ability is the Free Energy Principle (FEP), which posits that perception is a generative process where the brain minimises Variational Free Energy (VFE) to maintain accurate internal models of the world. While Deep Neural Networks (DNNs) have served as powerful analogues for biological brains, they typically lack the real-time plasticity required to handle abrupt sensory shifts. We introduce FEP-Nav, a biologically inspired framework for real-time perceptual adaptation in robust visual navigation. Motivated by the decomposition of VFE into prediction error and Bayesian surprise, FEP-Nav combines a Top-down Decoder, which provides an internal expectation of uncorrupted sensory input, with Adaptive Normalisation, which adjusts shifted feature distributions toward prior statistics. We interpret reconstruction and normalisation as approximate mechanisms for reducing the corresponding VFE-related terms during inference without gradient-based updates. Experiments across simulated and real-world visual corruptions show that FEP-Nav restores performance lost under visual corruption, outperforming non-adaptive baselines and strong adaptive methods. These results suggest that variational principles can provide a useful design perspective for robust autonomous behaviour under degraded sensory conditions.

\end{abstract}


\section{Introduction}

\IEEEPARstart{N}{avigation} is a fundamental capability of living organisms, from bacteria to primates. Humans can continue navigating despite sudden sensory degradation, such as raindrops obscuring their glasses. This raises a fundamental question: what mechanisms allow biological agents to maintain adaptive perception and action when facing noisy, corrupted sensory streams?

A prominent hypothesis in theoretical neuroscience states that perception is a generative process of constructing the most probable representations or embeddings $Z$ of the world. The representations $Z$ are learnt and updated by minimising \emph{Variational Free Energy (VFE)}, formalised within the normative framework known as the \emph{Free Energy Principle (FEP)} \cite{friston2010free, parr2022active}. \emph{VFE} essentially contains two terms: prediction error and Bayesian surprise. The first measures the discrepancy between top-down predictions and sensory input, whereas the second measures the divergence between the posterior representation and its prior.

\emph{Deep Neural Networks (DNNs)} have been used as analogues of biological brains at multiple levels, from computational goals to algorithms that construct representations. Previous studies explore \emph{DNNs} trained using the \emph{VFE} objectives to propose computational hypotheses of how brains might work \cite{ueltzhoffer2018deep, van2020deep, fountas2020deep, mazzaglia2021contrastive, meo2021multimodal}. However, they do not consider how these \emph{DNNs} can continuously update their predictions under real-time constraints and do not demonstrate their practicality for solving real-world tasks, especially when the sensory distribution shifts abruptly, requiring instantaneous model updates.

\begin{figure}[t]
\centering
\includegraphics[width=0.8\linewidth]{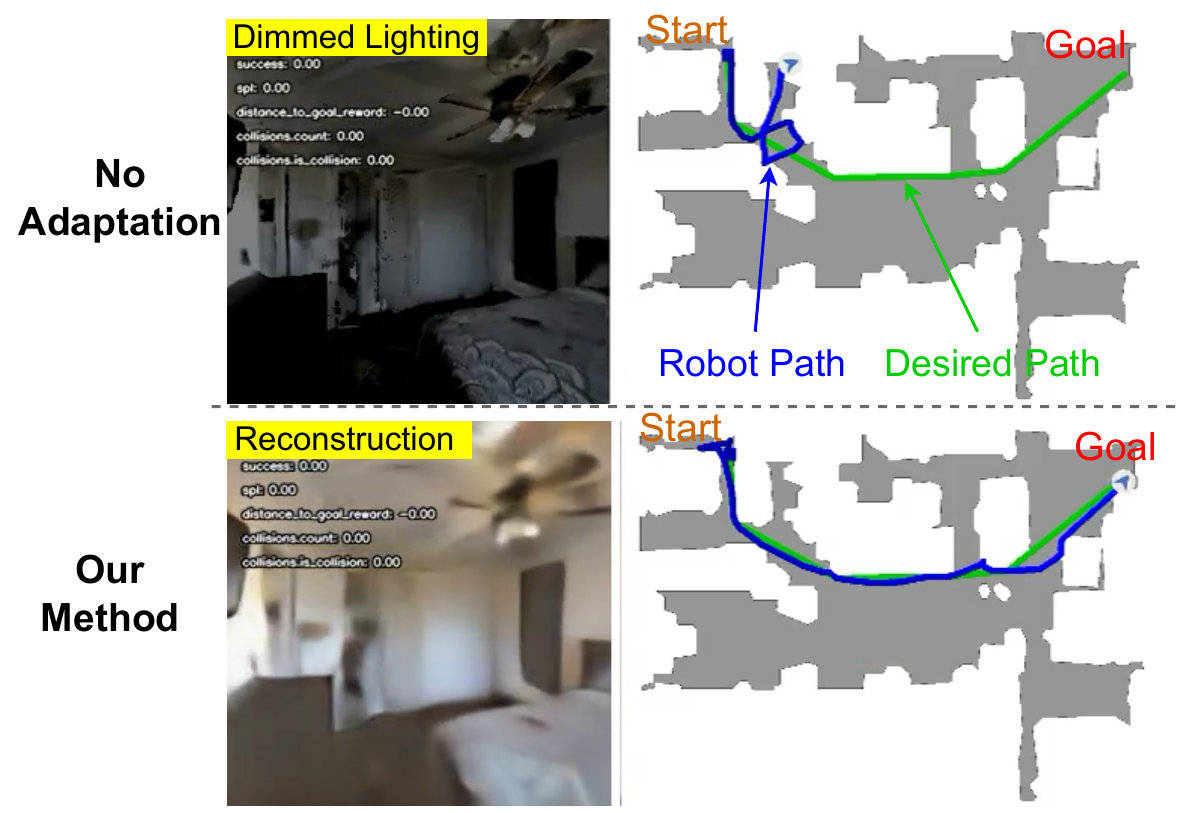}
\vspace{-3mm}
\caption{\textbf{Navigation under visual corruption:} \textit{Top:} A non-adaptive robot fails to reach the goal under dim lighting. \textit{Bottom:} FEP-Nav successfully reaches the goal.}
\vspace{-6mm}
\label{fig:intro}
\end{figure}

\begin{figure*}[!t]
\centering
\includegraphics[scale=0.35]{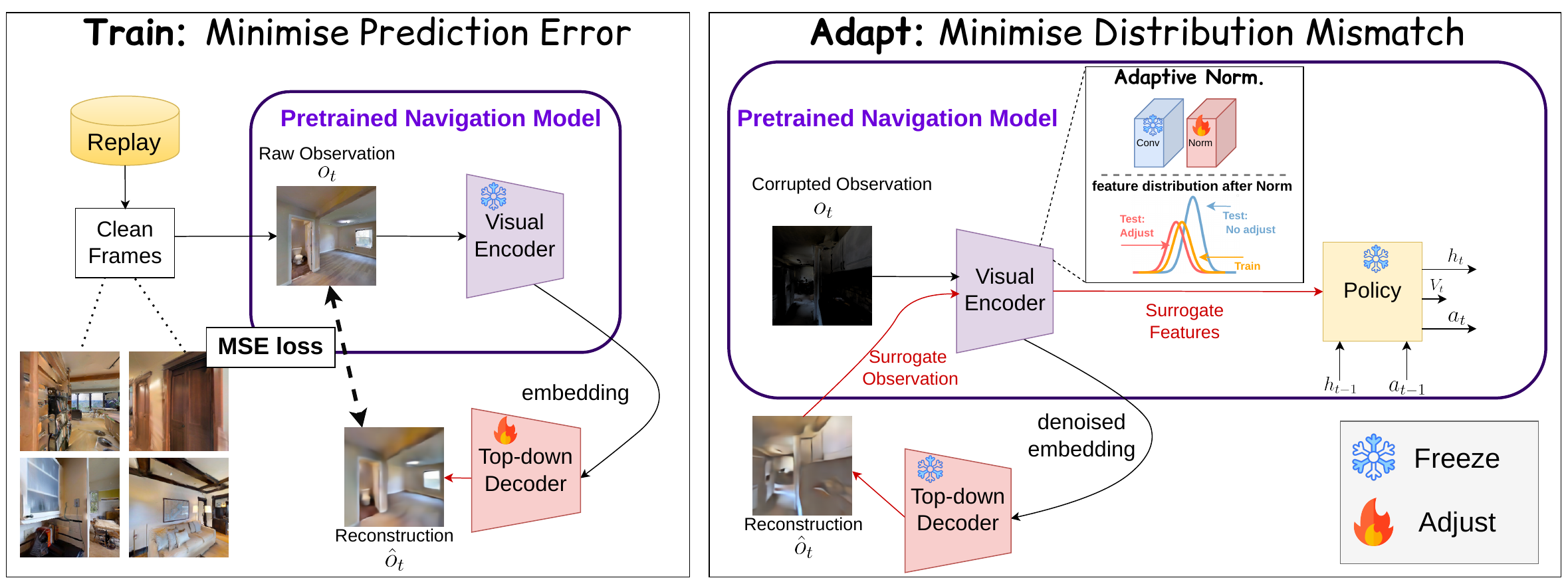}
\vspace{-2mm}
\caption{\textbf{An overview of our proposed method (FEP-Nav):} The pre-trained navigation agent comprises a Visual Encoder ($VE$) and a policy network. \textbf{Left (Offline Learning):} To minimise prediction error, a Top-down Decoder ($TD$) is trained via mean squared error (MSE) to reconstruct observations from $VE$'s extracted features, while the $VE$ remains frozen. Training utilises clean frames from a replay buffer of uncorrupted scenes. \textbf{Right (Online Adaptation):} To minimise feature-distribution mismatch during inference, the $TD$ remains frozen while the $VE$ BatchNorm layers dynamically update their normalisation statistics to counter sensory non-stationarity. The $TD$ processes these stabilised latent representations to output a clean surrogate observation, which is re-encoded by the $VE$ into surrogate features for the policy network to produce actions.
}
\vspace{-4mm}
\label{fig:ttanav}
\end{figure*}

To address this gap, we propose an \emph{FEP-inspired} method for real-time perceptual adaptation under visual uncertainty in both simulation and the real world. Consider a DNN-based perception model with low-level processing at earlier layers and high-level processing at later layers. Predictive-processing accounts associated with \emph{FEP} emphasise top-down signals that encode predictions of sensory input \cite{kreiman2020beyond,rao1999predictive, lotter2017deep}. Therefore, we propose a \emph{Top-down Decoder (TD)}, which receives the high-level features, reconstructs the inputs, and feeds reconstructed inputs back to the perception model. 

The perception model incorporates \emph{Adaptive Normalisation (AN)}, which can be interpreted as reducing a moment-based proxy for the Bayesian surprise term of \emph{VFE}. During test time, \emph{AN} adjusts normalisation statistics, denoises input features, and enables the decoder to reconstruct cleaner inputs. Our method enables the robot to navigate using these reconstructions as surrogate inputs instead of the actual ones. \emph{TD} is trained via offline self-supervised learning on uncorrupted inputs to learn representations capable of reconstructing clean observations. Importantly, the decoder can denoise the inputs during test time. The decoder output becomes the internal expectation of the world, allowing the robot to navigate based on what the clean input should look like, despite the corrupted input.

\begin{figure}[t]
\centering
\includegraphics[scale=0.45]{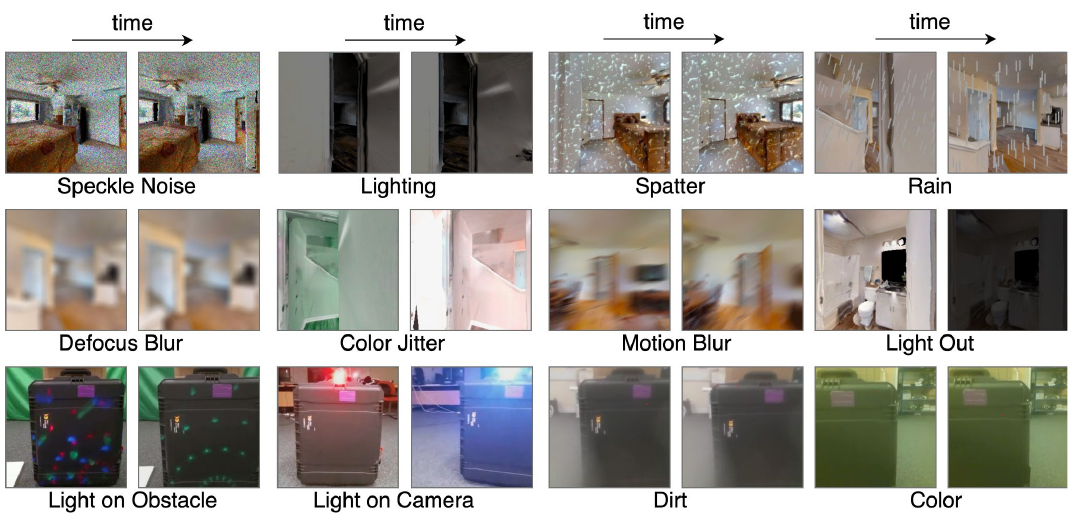}
\vspace{-4mm}
\caption{\textbf{Visual Corruptions:} Pairs of consecutive images of the same corruption type are presented in temporal order, as indicated by the arrow. The top and middle rows show simulated corruptions in the Habitat simulation, while the bottom row shows real-world corruptions.
}
\vspace{-3mm}
\label{fig:visual_corruption}
\end{figure}

To evaluate our proposed method, we created a navigation benchmark that includes eight simulated and four real visual corruptions to assess how navigation performance declines with each type of corruption, as shown in Fig. \ref{fig:visual_corruption}. We show that the strong \emph{DNN} baseline, trained with Reinforcement Learning (RL) on nearly 3 billion frames from diverse scenes, fails to navigate to the target position under trivial visual corruption (e.g., dimmed lighting), as shown in Fig.~\ref{fig:intro}, despite the fact that it achieves a near-perfect success rate on unseen indoor scenes without visual corruption. 

Our method substantially improves navigation performance over non-adaptive baselines, which cannot adjust visual representation during test time. Our method also outperforms five state-of-the-art adaptive methods, which update the visual representation during test time. Moreover, our lightweight method operates in real time on a physical robot and substantially restores navigation performance across four visual corruptions.

Our novel contributions are as follows: \\
\noindent \textbf{1.} We propose an \emph{FEP-inspired} real-time perceptual adaptation framework for robust visual navigation, and demonstrate its practicality in both simulated and real-world robotic navigation under visual corruptions. \\
\noindent \textbf{2.} We provide a theoretical interpretation showing how two standard deep learning mechanisms, normalisation and reconstruction, can be connected to the VFE-inspired notions of Bayesian surprise and prediction error, thereby offering insight into the relationship between machine learning, robotics, and neuroscience.

\section{Related Work}
\subsection{The Free Energy Principle in Robotics}
The \emph{Free Energy Principle (FEP)} in robotics is mainly implemented as the \emph{Active Inference Framework (AIF)}~\cite{parr2022active}, which unifies state estimation~\cite{bos2022free, meera2023adaptive}, control, and learning~\cite{baioumy2021active} into a single optimisation problem~\cite{da2022active, lanillos2021active}. As scaling these agents to high-dimensional state spaces requires \emph{DNNs}, recent studies leverage them to drive more complex behaviours~\cite{mazzaglia2021contrastive, meo2021multimodal, meo2022adaptation, wirkuttis2021leading}. Fountas et al.~\cite{fountas2020deep} mark a seminal advancement in this domain, showing the efficacy of deep active inference in high-dimensional state spaces. Similarly, Fujii and Murata~\cite{fujii2025real} introduce temporally hierarchical world models to enable computationally efficient planning, while Vijayaraghavan et al. develop linguistic and sensorimotor compositionality for generalising to novel tasks~\cite{vijayaraghavan2025development}.

While \emph{AIF} unifies perception and action, Lanillos and Cheng~\cite{lanillos2018adaptive} demonstrated the utility of pure \emph{Perceptual Inference} for robust state estimation, minimising \emph{VFE} only to update internal beliefs. Adopting this stance, we use VFE-inspired principles to guide test-time representation adaptation, without performing explicit variational inference or exact \emph{VFE} minimisation. This decoupling enables us to leverage pretrained high-performance policies from Reinforcement Learning (RL) and Imitation Learning (IL), avoiding the complexity in training deep \emph{AIF} agents from scratch. Importantly, a distinct feature of our method is its reliance on only the feedforward operation of the \emph{DNNs} without gradient updates, enabling light-weight, real-time adaptation.

\subsection{Test-time Adaptation}
Adapting a model to distribution shifts using incoming test data is known as test-time adaptation (TTA), enabling models to adapt to unknown distributions during inference~\cite{wang2020tent,liang2020shot,mirza2022dua,lin2023vitta}. Despite its success in other vision domains~\cite{wang2023diga, lin2023vitta}, real-time TTA for visual corruption in navigation is under-explored, as most SOTA methods are task-specific (e.g., assuming categorical outputs) and require fine-tuning time on the test batch. Previous studies also introduce TTA in robot navigation to improve a robot's trajectories in novel simulated scenes without visual corruption, but their methods are not real-time and require fine-tuning that needs data collection from a new test episode \cite{ko2025active, kim2025test}. Our Adaptive Normalisation is related to normalisation-based TTA methods such as AdaBN, DUA, and MemBN, which update BatchNorm statistics at inference time~\cite{li2016revisiting, mirza2022dua,kang2024membn}. The main contribution of FEP-Nav is to combine this lightweight adaptation with a Top-down Decoder, producing surrogate clean observations and providing an FEP-inspired interpretation through prediction-error and Bayesian-surprise reduction. FEP-Nav outperforms five TTA methods~\cite{wang2020tent, liang2020shot, mirza2022dua, kang2024membn, kim2026buffer} and works in real-time robot navigation.

\section{Methods}
\label{sec:methods}
This section explains the problem formulation of the point-goal navigation and the implementation of our proposed method. We then provide a theoretical link between our proposed method and Free-Energy Principle (FEP).
\subsection{Point-Goal Navigation Problem Formulation}

In point-goal navigation, a robot is tasked with moving to the given relative direction: $(\Delta x_{\text{target}}, \; \Delta y_{\text{target}}) \triangleq (x_{\text{target}}-x_{\text{start}}, y_{\text{target}}-y_{\text{start}})$ in a cluttered indoor scene. 
For instance, given the goal of moving to $(\Delta x_{\text{target}} = 0.5m, \Delta y_{\text{target}}=-2m)$, 
the robot has to move to the target position $(x_{\text{start}} + 0.5, y_{\text{start}}-2)$, where $(x_{\text{start}}, y_{\text{start}})$ is the initial position. A shorter path leads to a higher reward.
In our setting, the robot is equipped with a navigation model trained with clean RGB observations and GPS \& Compass $(x_{\text{target}}-x_{t}, y_{\text{target}}-y_{t})$. The GPS  \& Compass is derived from the error-free odometry $(\Delta x_{t}, \Delta y_{t})$. Where $\Delta x_{t}= x_{t} - x_{t-1}$, $\Delta y_{t}= y_{t} - y_{t-1}$, and ($x_t$, $y_t$) is the position of the robot at time $t$.

\subsection{End-to-End Visual Navigation Models}
\label{subsec:policy_training}
An end-to-end visual navigation model typically consists of a Visual Encoder (VE) and a policy network. The VE extracts features from RGB observations, while the recurrent policy network uses these features together with the target location to predict actions~\cite{zhu2017rl_nav}. The architecture used in this work is shown in the purple frame of \autoref{fig:ttanav}.

We adopt the DD-PPO point-goal navigation model~\cite{wijmans2019ddppo}, trained on 2.5 billion RGB frames from the Gibson $2+$ training scenes~\cite{xiazamirhe2018gibsonenv}. Since our method adapts BatchNorm statistics at test time, but the original DD-PPO uses GroupNorm~\cite{wu2018groupnrom}, we replace all GroupNorm layers with BatchNorm and fine-tune the model on Gibson scenes for 120M frames using the same training configuration and reward function as~\cite{wijmans2019ddppo}. The VE uses SE-ResNeXt-50~\cite{hu2018squeeze}, and the policy network is an LSTM with hidden size $512$. We refer to this BatchNorm-based model as \emph{Pretrained-Nav}.

\subsection{FEP-Nav: Implementation}
\vspace{-3mm}
\label{subsec:proposed_method}
\begin{algorithm}[h]
	\caption{FEP-Nav Training and Inference}
	\label{algo:ttanav_full}
    \small
	\begin{algorithmic}[1]
        \State \textbf{Definitions:} \emph{VE} $f_{\phi}$; \emph{TD} $g_{\theta}$; Policy $\pi_{\psi}$; LSTM hidden state $h_t$; block3 output $f^{b_{3}}_{\phi}(\cdot)$.
        
		\Procedure{OfflineTraining}{Dataset}
            \State Freeze VE parameters $\phi$
    		\While {not converged}
        		\State $o \gets \text{Sample}(\text{Dataset})$
        		\State $Z \gets f^{b_{3}}_{\phi}(o)$
        		\State $\hat{o} \gets g_{\theta}(Z)$
        		\State Update $\theta$ via Adam minimising $\sum (o - \hat{o})^{2}$
    		\EndWhile
		\EndProcedure
        
		\Procedure{OnlineInference}{Env, Target}
            \State Enable updates for $\mu_{k},\sigma_{k}$ in $f_{\phi}$ (Eq. \ref{eq:running_mean}, \ref{eq:running_var})
    		\For {each time step $t$}
        		\State $o_{t} \gets \text{Env.GetObs}()$
        		\State $Z_{t} \gets f^{b_{3}}_{\phi}(o_{t})$ \Comment{Update and use adapted $\hat{\mu},\hat{\sigma}$}
        		\State $\hat{o}_{t} \gets g_{\theta}(Z_{t})$ \Comment{Reconstruct clean observation}
        		\State $e_{t} \gets f_{\phi}(\hat{o}_{t})$ \Comment{Re-encode using original training $\mu,\sigma$}
        		\State $a_{t}, h_t \gets \pi_{\psi}(e_{t}, h_{t-1}, \text{Target})$
        		\State Env.Step($a_t$)
    		\EndFor
		\EndProcedure
	\end{algorithmic}
\end{algorithm}

FEP-Nav introduces \emph{Top-down Decoder (TD)} denoted as $g_{\theta}$ that receives the output of the late layer of VE and predicts the reconstructed image $\hat{o_{t}}$. See Fig. \ref{fig:ttanav}. \emph{TD} comprises residual block modules adapted from architectures proposed in~\cite{child2020very,pandey2022diffusevae}. This decoder produces RGB images with dimensions of $256 \times 256$. It has 5.3M trainable parameters. The overall process is described in \textbf{Algorithm~\ref{algo:ttanav_full}}.

\textbf{Training:} 
During the training of FEP-Nav, we freeze all VE parameters and train \emph{TD} using mean squared error (MSE) reconstruction loss with samples from 72 training scenes of Gibson \cite{xiazamirhe2018gibsonenv} without visual corruption. 
\emph{TD} is trained with a very small subset of navigation data: 112k out of 2.5 billion frames. We follow the same optimisation hyperparameters as in \cite{pandey2022diffusevae}. Specifically, we use Adam optimiser with the learning rate of $2 \times 10^{-5}$, and the momentum of $0.9999$.

\textbf{Adaptation:} 
FEP-Nav allows statistics in the BatchNorm layers of the VE to be updated during test time. We call this Adaptive Normalisation (AN). BatchNorm layers normalise each feature map (channel) separately, meaning that pixels in a feature map are normalised by the same estimated moments. Suppose we have a CNN layer with a set of feature maps $Z = [Z^{1}, \dots,Z^{C}] \in \mathcal{R}^{H \times W \times C}$ where $H \times W$ denote width and height and $C$ is the number of feature maps. For simplicity, we consider each pixel $z$ of a feature map $Z^{i}$, where $i \in \{1,\dots,C\}$. Specifically, each batch normalisation layer normalises an incoming sample $z$ by the estimated mean and variance as in \eqref{eq:batchnorm_layer},
\begin{equation}
\label{eq:batchnorm_layer}
    z_{\text{norm}} = \frac{z-\hat{\mu}_{k}}{\hat{\sigma}_{k}}\cdot\gamma + \beta,
\end{equation}
where $\gamma$ and $\beta$ are affine parameters that scale and shift the data distribution respectively. 
Instead of estimating the mean and variance by averaging samples in a batch, we set VE to update the mean and variance by the moving average equations as in \eqref{eq:running_mean} and \eqref{eq:running_var} below,
\begin{align}
\hat{\mu}_{k} &= (1-\rho)\cdot\hat{\mu}_{k-1} + \rho\cdot\mu_{k},  \label{eq:running_mean} \\
\hat{\sigma}^{2}_{k} &= (1-\rho)\cdot\hat{\sigma}^{2}_{k-1} + \rho\cdot\sigma^{2}_{k},  \label{eq:running_var}
\end{align}
where $\hat{\mu}_{k}$ is the estimated mean and $\hat{\sigma}^{2}_{k}$ is the estimated variance, 
$\mu_{k}$ and $\sigma^{2}_{k}$ are the sample mean and variance of the batch data, 
and $\rho$ is a hyperparameter called momentum, which determines the contribution of the most recent sample estimate ($\mu_{k},\sigma^{2}_{k}$) to the calculation of the average. 
Here we update $\hat{\mu}_{k}$ and $\hat{\sigma}^{2}_{k}$ every inference step so $\mu_{k}$ and $\sigma^{2}_{k}$ are computed from instance statistics. Our model can adapt to the visual change at every action step and does not have to wait until it has enough samples in the batch, which may be too late for the robot to adjust its behaviour. Finally, the pretrained decoder, acting as the model’s internal expectation of the world, uses the re-normalised representation to predict the image. The decoder remains frozen and is trained to infer the expected appearance based on the given representation. We introduce two variants of FEP-Nav: FEP-Nav Running and FEP-Nav Instance. FEP-Nav Running updates the statistics as in \eqref{eq:running_mean} and \eqref{eq:running_var}, where adaptive $\rho = f(t) < 1$, while FEP-Nav Instance only uses instance statistics by setting $\rho = 1$. In other words, FEP-Nav Instance reduces moving average to instance normalisation.

\subsection{Theoretical Motivation}
\label{subsec:probabilistic_formulation}

We interpret FEP-Nav through the lens of the Free Energy Principle (FEP) \cite{friston2010free, baioumy2022unbiased}. Under the VFE formulation, perception can be associated with two terms: prediction error and Bayesian surprise:
\begin{equation}
\label{eq:elbo_decomposition}
\mathcal{L}_{\text{VFE}} = \underbrace{\mathbb{E}_{q_\phi(z|o)}[-\ln p_\theta(o|Z)]}_{\text{Prediction Error}} + \underbrace{D_{KL}[q_\phi(Z|o) \| p(Z)]}_{\text{Bayesian Surprise}}
\end{equation}
Here, the implemented algorithm uses deterministic reconstruction and feature normalisation as practical proxies for the two VFE-related terms. In FEP-Nav, \emph{TD} is trained to reduce a reconstruction-based proxy for prediction error, while \emph{AN} can be interpreted as reducing a moment-based proxy for Bayesian surprise during test time. To make this interpretation tractable, we adopt two standard approximations:

\newtheorem{assumption}{Assumption}

\begin{assumption}[Mean-Field Approximation]
\label{asm:mean_field}
We approximate the complex posterior distribution $q(Z|o)$ with a factorised variational density. Specifically, we treat the feature at distinct spatial locations $i \in \{1, \dots, N\}$ as independent in the variational posterior:
\begin{equation}
q(Z|o) \approx \prod_{i=1}^{N} q(z^{(i)}|o)
\end{equation}
This standard approximation makes the ELBO tractable.
\end{assumption}

\begin{assumption}[Spatial Ergodicity]
\label{asm:ergodicity}
The feature process is spatially ergodic \cite{ulyanov2016instance}. This implies that the spatial average over a single realisation converges in probability to the ensemble expectation as the spatial dimensions $N$ grow:
\begin{align}
\hat{\mu} &= \frac{1}{N}\sum_{i=1}^{N} z^{(i)} \xrightarrow{P} \mathbb{E}[z] \\
\hat{\sigma}^2 &= \frac{1}{N}\sum_{i=1}^{N} (z^{(i)} - \hat{\mu})^2 \xrightarrow{P} \text{Var}(z)
\end{align}
This property justifies the use of estimated moments as proxies for the true ones.
\end{assumption}
AN can be viewed as moment matching in latent space: visual corruptions shift feature means and variances, while test-time normalisation re-centres and re-scales them. Under a Gaussian approximation, aligning these moments reduces the KL divergence to the assumed prior, linking AN to Bayesian-surprise reduction.
\begin{proposition}
\label{thm:reduce_kl}
Let $p(z) = \mathcal{N}(0, I)$ be the prior. Let the corrupted feature distribution be $q(z|o) = \mathcal{N}(\mu_{corr}, \sigma_{corr}^2)$, where the corruption induces a significant distribution shift $\Delta$ away from the prior parameters $(0,1)$.
Let the normalised distribution be $q(z_{norm})$, computed using empirical statistics $\hat{\mu}$ and $\hat{\sigma}$.
Given that the estimation error of the statistics is smaller than the distribution shift caused by corruption ($\epsilon \ll \Delta$), the KL divergence decreases:
\begin{equation}
D_{KL}(q(z_{norm}|\cdot) \| p(z)) < D_{KL}(q(z|\cdot) \| p(z))
\end{equation}
\end{proposition}

\textbf{Scope of the analysis.}
The above result provides a principled interpretation of AN as a mechanism for reducing Bayesian surprise under feature-distribution shift. In particular, when corrupted observations shift the latent feature moments away from the training-time prior, normalising features using updated test-time statistics moves the adapted representation closer to the assumed prior in moment space. This yields a reduction in the corresponding KL term under the Gaussian, mean-field, and spatial-ergodicity approximations used in the analysis. While these approximations abstract away from the full complexity of CNN feature distributions, they provide a tractable explanation for why AN can support VFE-inspired perceptual adaptation in real time. We further support this interpretation empirically by measuring feature-space KL divergence before and after adaptation in Sec.~\ref{sec:experiment} (\autoref{tab:kl_summary_pre_lstm}).

\vspace{-2mm}

\begin{table}[ht]
\centering
\caption{Empirical KL divergence ($\downarrow$) between corrupted and clean latent features with and without adaptation. Values represent mean $\pm$ SD.}
\vspace{-3mm}
\label{tab:kl_summary_pre_lstm}
\resizebox{0.5\columnwidth}{!}
{
\begin{tabular}{lcc}
\toprule
Corruption Types & \multicolumn{1}{c}{Without AN} & \multicolumn{1}{c}{With AN} \\
\midrule
Defocus Blur & 54.3 $\pm$ 31.5 & \textbf{46.9 $\pm$ 30.8} \\
Jitter & 27.7 $\pm$ 3.6 & \textbf{13.7 $\pm$ 3.7} \\
Light Out & 103.2 $\pm$ 10.5 & \textbf{32.4 $\pm$ 24.0} \\
Lighting & 87.2 $\pm$ 33.5 & \textbf{48.9 $\pm$ 30.4} \\
Motion Blur & \textbf{20.6 $\pm$ 4.7} & 20.7 $\pm$ 3.8 \\
Rain & 139.9 $\pm$ 17.1 & \textbf{23.2 $\pm$ 2.1} \\
Spatter & 35.4 $\pm$ 4.2 & \textbf{16.1 $\pm$ 1.9} \\
Speckle Noise & 40.2 $\pm$ 5.7 & \textbf{15.7 $\pm$ 2.1} \\
\bottomrule
\end{tabular}
}
\vspace{-5mm}
\end{table}

\begin{table*}[t]
\vspace{+0.3cm}
\centering
\caption{\textbf{Navigation performance under corruptions.} Entries show mean $\pm$ SD over evaluation scenes. $*$ indicates severe DD-PPO degradation. \textbf{Bold}/\textbf{\textcolor[HTML]{D17A22}{brown}} mean best/second-best per row. See Sec.~\ref{sec:experiment}.}
\vspace{-0.2cm}
\label{table:performance}
\resizebox{2.00\columnwidth}{!}
{
\begin{tabular}{lcccccccccccccccccc}
\toprule
\multirow{2.5}{*}{Corruption Type $\downarrow$} & \multicolumn{2}{c}{DD-PPO} & \multicolumn{2}{c}{Pretrained-Nav} & \multicolumn{2}{c}{TENT} & \multicolumn{2}{c}{SHOT-IM} & \multicolumn{2}{c}{DUA} & \multicolumn{2}{c}{MemBN} & \multicolumn{2}{c}{Buffer} & \multicolumn{2}{c}{FEP-Nav Running} & \multicolumn{2}{c}{FEP-Nav Instance} \\
\cmidrule(lr){2-3} \cmidrule(lr){4-5} \cmidrule(lr){6-7} \cmidrule(lr){8-9} \cmidrule(lr){10-11} \cmidrule(lr){12-13} \cmidrule(lr){14-15} \cmidrule(lr){16-17} \cmidrule(lr){18-19}
 & SR & SPL & SR & SPL & SR & SPL & SR & SPL & SR & SPL & SR & SPL & SR & SPL & SR & SPL & SR & SPL \\
\midrule
Clean & \textbf{0.98 $\pm$ 0.05} & \textbf{0.91 $\pm$ 0.08} & 0.96 $\pm$ 0.08 & \textcolor[HTML]{D17A22}{\textbf{0.88 $\pm$ 0.10}} & 0.94 $\pm$ 0.08 & 0.82 $\pm$ 0.12 & 0.94 $\pm$ 0.09 & 0.82 $\pm$ 0.12 & 0.96 $\pm$ 0.07 & 0.85 $\pm$ 0.12 & 0.96 $\pm$ 0.07 & 0.85 $\pm$ 0.11 & \textcolor[HTML]{D17A22}{\textbf{0.97 $\pm$ 0.08}} & 0.88 $\pm$ 0.10 & 0.96 $\pm$ 0.06 & 0.87 $\pm$ 0.09 & 0.96 $\pm$ 0.07 & 0.86 $\pm$ 0.10 \\
Speckle Noise* & 0.48 $\pm$ 0.20 & 0.32 $\pm$ 0.16 & 0.67 $\pm$ 0.18 & 0.40 $\pm$ 0.14 & 0.89 $\pm$ 0.14 & 0.76 $\pm$ 0.16 & 0.91 $\pm$ 0.12 & 0.77 $\pm$ 0.14 & 0.94 $\pm$ 0.10 & \textbf{0.82 $\pm$ 0.14} & 0.92 $\pm$ 0.11 & 0.78 $\pm$ 0.14 & 0.65 $\pm$ 0.20 & 0.40 $\pm$ 0.17 & \textbf{0.95 $\pm$ 0.09} & \textcolor[HTML]{D17A22}{\textbf{0.81 $\pm$ 0.12}} & \textcolor[HTML]{D17A22}{\textbf{0.94 $\pm$ 0.09}} & 0.81 $\pm$ 0.13 \\
Lighting* & 0.52 $\pm$ 0.23 & 0.30 $\pm$ 0.18 & 0.70 $\pm$ 0.22 & 0.44 $\pm$ 0.17 & 0.86 $\pm$ 0.17 & 0.68 $\pm$ 0.19 & 0.86 $\pm$ 0.17 & 0.68 $\pm$ 0.19 & 0.92 $\pm$ 0.12 & 0.76 $\pm$ 0.16 & 0.87 $\pm$ 0.17 & 0.69 $\pm$ 0.19 & 0.68 $\pm$ 0.24 & 0.43 $\pm$ 0.19 & \textbf{0.92 $\pm$ 0.12} & \textbf{0.80 $\pm$ 0.16} & \textcolor[HTML]{D17A22}{\textbf{0.92 $\pm$ 0.12}} & \textcolor[HTML]{D17A22}{\textbf{0.79 $\pm$ 0.17}} \\
Spatter* & 0.60 $\pm$ 0.22 & 0.42 $\pm$ 0.18 & 0.71 $\pm$ 0.19 & 0.45 $\pm$ 0.14 & 0.90 $\pm$ 0.13 & 0.74 $\pm$ 0.15 & 0.89 $\pm$ 0.15 & 0.75 $\pm$ 0.15 & 0.93 $\pm$ 0.11 & 0.80 $\pm$ 0.14 & 0.91 $\pm$ 0.12 & 0.76 $\pm$ 0.15 & 0.69 $\pm$ 0.19 & 0.45 $\pm$ 0.15 & \textcolor[HTML]{D17A22}{\textbf{0.95 $\pm$ 0.07}} & \textbf{0.84 $\pm$ 0.11} & \textbf{0.96 $\pm$ 0.07} & \textcolor[HTML]{D17A22}{\textbf{0.83 $\pm$ 0.11}} \\
Rain* & 0.04 $\pm$ 0.04 & 0.02 $\pm$ 0.02 & 0.23 $\pm$ 0.15 & 0.08 $\pm$ 0.06 & 0.49 $\pm$ 0.23 & 0.23 $\pm$ 0.13 & 0.52 $\pm$ 0.25 & 0.25 $\pm$ 0.14 & \textcolor[HTML]{D17A22}{\textbf{0.83 $\pm$ 0.20}} & \textcolor[HTML]{D17A22}{\textbf{0.62 $\pm$ 0.18}} & 0.43 $\pm$ 0.22 & 0.20 $\pm$ 0.11 & 0.21 $\pm$ 0.15 & 0.07 $\pm$ 0.05 & 0.58 $\pm$ 0.24 & 0.32 $\pm$ 0.15 & \textbf{0.89 $\pm$ 0.16} & \textbf{0.73 $\pm$ 0.18} \\
Defocus Blur* & 0.70 $\pm$ 0.25 & 0.56 $\pm$ 0.24 & 0.75 $\pm$ 0.20 & 0.53 $\pm$ 0.20 & 0.86 $\pm$ 0.17 & 0.70 $\pm$ 0.17 & 0.87 $\pm$ 0.15 & 0.70 $\pm$ 0.16 & \textbf{0.92 $\pm$ 0.11} & \textbf{0.79 $\pm$ 0.15} & \textcolor[HTML]{D17A22}{\textbf{0.88 $\pm$ 0.17}} & \textcolor[HTML]{D17A22}{\textbf{0.72 $\pm$ 0.18}} & 0.75 $\pm$ 0.21 & 0.54 $\pm$ 0.20 & 0.85 $\pm$ 0.18 & 0.66 $\pm$ 0.19 & 0.83 $\pm$ 0.21 & 0.67 $\pm$ 0.20 \\
Motion Blur & 0.82 $\pm$ 0.22 & 0.70 $\pm$ 0.24 & 0.81 $\pm$ 0.23 & 0.63 $\pm$ 0.22 & 0.88 $\pm$ 0.14 & 0.73 $\pm$ 0.16 & 0.89 $\pm$ 0.13 & 0.74 $\pm$ 0.15 & \textbf{0.93 $\pm$ 0.12} & \textbf{0.79 $\pm$ 0.15} & \textcolor[HTML]{D17A22}{\textbf{0.90 $\pm$ 0.15}} & \textcolor[HTML]{D17A22}{\textbf{0.75 $\pm$ 0.18}} & 0.82 $\pm$ 0.22 & 0.63 $\pm$ 0.22 & 0.84 $\pm$ 0.21 & 0.67 $\pm$ 0.22 & 0.86 $\pm$ 0.20 & 0.69 $\pm$ 0.21 \\
Colour Jitter & 0.91 $\pm$ 0.14 & 0.78 $\pm$ 0.18 & 0.81 $\pm$ 0.20 & 0.61 $\pm$ 0.20 & 0.90 $\pm$ 0.14 & 0.77 $\pm$ 0.17 & \textcolor[HTML]{D17A22}{\textbf{0.92 $\pm$ 0.10}} & 0.79 $\pm$ 0.14 & 0.81 $\pm$ 0.19 & 0.63 $\pm$ 0.19 & 0.92 $\pm$ 0.11 & \textcolor[HTML]{D17A22}{\textbf{0.80 $\pm$ 0.15}} & 0.80 $\pm$ 0.22 & 0.60 $\pm$ 0.20 & 0.80 $\pm$ 0.19 & 0.60 $\pm$ 0.19 & \textbf{0.95 $\pm$ 0.08} & \textbf{0.83 $\pm$ 0.13} \\
Light Out & 0.91 $\pm$ 0.10 & \textcolor[HTML]{D17A22}{\textbf{0.81 $\pm$ 0.15}} & 0.91 $\pm$ 0.13 & 0.78 $\pm$ 0.15 & 0.92 $\pm$ 0.11 & 0.78 $\pm$ 0.15 & 0.91 $\pm$ 0.13 & 0.78 $\pm$ 0.15 & 0.76 $\pm$ 0.20 & 0.51 $\pm$ 0.17 & \textcolor[HTML]{D17A22}{\textbf{0.92 $\pm$ 0.11}} & 0.78 $\pm$ 0.15 & 0.92 $\pm$ 0.12 & 0.78 $\pm$ 0.15 & 0.77 $\pm$ 0.20 & 0.52 $\pm$ 0.19 & \textbf{0.97 $\pm$ 0.06} & \textbf{0.86 $\pm$ 0.09} \\
\midrule
\textbf{Average} & 0.66 $\pm$ 0.30 & 0.54 $\pm$ 0.29 & 0.73 $\pm$ 0.21 & 0.53 $\pm$ 0.23 & 0.85 $\pm$ 0.14 & 0.69 $\pm$ 0.18 & 0.86 $\pm$ 0.13 & 0.70 $\pm$ 0.17 & \textcolor[HTML]{D17A22}{\textbf{0.89 $\pm$ 0.07}} & \textcolor[HTML]{D17A22}{\textbf{0.73 $\pm$ 0.11}} & 0.86 $\pm$ 0.16 & 0.70 $\pm$ 0.20 & 0.72 $\pm$ 0.22 & 0.53 $\pm$ 0.23 & 0.85 $\pm$ 0.12 & 0.68 $\pm$ 0.18 & \textbf{0.92 $\pm$ 0.05} & \textbf{0.79 $\pm$ 0.07} \\
\textbf{Minimum}
& 0.04 $\pm$ 0.04
& 0.02 $\pm$ 0.02
& 0.23 $\pm$ 0.15
& 0.08 $\pm$ 0.06
& 0.49 $\pm$ 0.23
& 0.23 $\pm$ 0.13
& 0.52 $\pm$ 0.25
& 0.25 $\pm$ 0.14
& \textcolor[HTML]{D17A22}{\textbf{0.76 $\pm$ 0.20}}
& \textcolor[HTML]{D17A22}{\textbf{0.51 $\pm$ 0.17}}
& 0.43 $\pm$ 0.22
& 0.20 $\pm$ 0.11
& 0.21 $\pm$ 0.15
& 0.07 $\pm$ 0.05
& 0.58 $\pm$ 0.24
& 0.32 $\pm$ 0.15
& \textbf{0.83 $\pm$ 0.21}
& \textbf{0.67 $\pm$ 0.20} \\
\bottomrule
\end{tabular}%
}
\vspace{-4mm}
\end{table*}

\begin{table}[ht]

\caption{Obstacle-avoidance success rates for the real-world robot under different adaptation methods. Each method is evaluated 30 episodes per corruption type.}
\vspace{-2mm}
\centering
\label{tab:drone_results}
\resizebox{0.7\columnwidth}{!}
{
\begin{tabular}{cccc}
 \toprule
Corruption Types $\downarrow$ & Vanilla Policy & DUA                                  & \begin{tabular}[c]{@{}c@{}}FEP-Nav\\ Instance (Ours)\end{tabular} \\ \midrule
Clean             & \textbf{0.9}      & 0.8                                  & {\color[HTML]{D17A22} \textbf{0.87}}                              \\
Colour             & 0.07              & {\color[HTML]{D17A22} \textbf{0.4}}  & \textbf{0.57}                                                     \\
Light on Camera   & 0.27              & {\color[HTML]{D17A22} \textbf{0.53}} & \textbf{0.57}                                                     \\
Dirt              & 0.27              & {\color[HTML]{D17A22} \textbf{0.3}}  & \textbf{0.8}                                                      \\
Light on Obstacle & 0.47              & {\color[HTML]{D17A22} \textbf{0.67}} & \textbf{0.77}                                                     \\ \midrule
\textbf{Average}  & 0.40              & {\color[HTML]{D17A22} \textbf{0.54}} & \textbf{0.72}                                                     \\
\textbf{Minimum}  & 0.07              & {\color[HTML]{D17A22} \textbf{0.3}}  & \textbf{0.57}               \\ \bottomrule                                      
\end{tabular}
}
\vspace{-3mm}
\end{table}

\section{Experiments}
\label{sec:experiment}

We first delineate the experimental setup, comprising visual corruptions, baseline methods, and evaluation metrics. The results are subsequently presented in the following sequence: (i) an analysis of the internal expectations reconstructed by the \emph{TD}; (ii) a performance benchmark against adaptive baselines; (iii) a comparison between our generative approach and image restoration models; (iv) validation of real-time performance on a physical robot; and (v) an ablation study concerning hierarchical normalisation. Notably, all models are trained exclusively on clean observations to evaluate zero-shot adaptation to out-of-distribution shifts.
\subsection{Visual Corruptions}
The implementations of \textit{Defocus Blur}, \textit{Motion Blur}, \textit{Lighting}, \textit{Spatter}, and \textit{Speckle Noise} are from the Habitat robustness benchmark dataset \cite{rajivc2022robustness}. The rest of the corruption types are implemented in the following ways: \textit{Light Out} is implemented by reducing the brightness of each frame with a $50\%$ probability, for example, approximately 50 of 100 sampled frames will be dark. \textit{Colour Jitter} is implemented by randomly adjusting the brightness, contrast, saturation, and hue of a frame. Real-world visual corruptions are implemented as follows. \textit{Light on Obstacle} is achieved by projecting disco lights onto a black suitcase. \textit{Light on Camera} is achieved by projecting disco lights onto a camera. \textit{Dirt} is implemented by applying dirt to a camera. \textit{Colour} is implemented by covering a camera with colour films.

\subsection{Baselines for Simulation Test}
We compare the proposed FEP-Nav methods with seven baselines.
Among these baselines, TENT, SHOT-IM, DUA, MemBN and Buffer are task-agnostic and can be seamlessly integrated into our navigation model. All baselines are evaluated with fourteen unseen scenes and 994 episodes \cite{xiazamirhe2018gibsonenv} used in \cite{mezghan2022memory}. We use Habitat simulator~\cite{habitat19iccv} where the robot has 4 actions: move forward (0.25m), turn left (10°), turn right (10°) and stop. An episode ends when the robot takes a stop action. During the evaluation, we allow the robot to take up to 500 environment steps in an episode. Parameters of DD-PPO and Pretrained-Nav are frozen during testing, while others have adaptation. 
\textbf{Pretrained-Nav} is our navigation model that is fine-tuned from DD-PPO as explained in \autoref{sec:methods}. 
The model is trained with 72 training scenes as in \cite{mezghan2022memory}. 
\textbf{DD-PPO \cite{wijmans2019ddppo}} is the SOTA on a point-goal navigation task. See \autoref{sec:methods} for more details. 
\textbf{DUA} \cite{mirza2022dua} has the same model architecture as Pretrained-Nav, with the difference that DUA updates the BatchNorm statistics $\hat\mu_{k}$ and $\hat\sigma_{k}$ during test time with running mean and variance as explained in \eqref{eq:running_mean} and \eqref{eq:running_var}.
\textbf{TENT \cite{wang2020tent}} is initialised with Pretrained-Nav's architecture and parameters. The model is then fine-tuned with entropy loss during test time. We follow the original implementation of TENT.
\textbf{SHOT-IM} \cite{liang2020shot} is initialised with the architecture and parameters of Pretrained-Nav. During test time, the model undergoes fine-tuning using the Information Maximisation (IM) loss.  \textbf{MemBN} \cite{kang2024membn} and \textbf{Buffer} \cite{kim2026buffer} introduce additional test-time adaptation layers that can be integrated with BatchNorm layers; therefore, we apply these methods to Pretrained-Nav. Buffer uses the TENT update rule and is therefore treated as a gradient-based adaptation method. Accordingly, for TENT, SHOT-IM, and Buffer, we reset the adapted parameters to their pretrained values at the beginning of each episode, as we find that accumulating updates across many episodes leads to model collapse.

\vspace{-2mm}
\subsection{Evaluation Metrics for Simulation Test}

The evaluation metrics below are used for the methods tested in the Habitat environment.
\textbf{Success Rate ($SR$)} measures the proportion of successful episodes among all evaluation episodes. An episode is deemed to be successful if an agent takes the stop action within $0.2m$ distance to a target position.

\textbf{Success weighted by Path Length ($SPL$)} is proposed in~\cite{anderson2018evaluation}. We follow the same definition.
Particularly, $SPL= S \frac{l}{\max(l,p)}$, where $S$ is a binary indicator of success in an episode.
$l$ is the shortest path between the start and target of an episode, and $p$ is the length of a path the agent takes.
$SPL$ is a strict metric in the sense that in order to achieve $SPL=1$ the agent must make no mistake, successfully arrive at the destination, and follow one of the shortest paths.

Regarding the last two rows of \autoref{table:performance}, \textbf{Average} is the average of SR or SPL over all corruptions. \textbf{Minimum} is the minimum SR or SPL over all corruptions, delineating the lower bound of navigation performance.

\begin{table}[b]
\vspace{-5mm}
\centering
\caption{Performance Comparison: Image Restoration vs. our method.}
\label{tab:restoration_comparison}
\resizebox{0.7\columnwidth}{!}
{
\begin{tabular}{lcccc}
\toprule
\multirow{2}{*}{Corruption Type} & \multicolumn{2}{c}{MPRNet} & \multicolumn{2}{c}{FEP-Nav (Ours)} \\
\cmidrule(lr){2-3} \cmidrule(lr){4-5}
 & SR & SPL & SR & SPL \\
\midrule
Defocus Blur & \textbf{0.92 $\pm$ 0.13} & \textbf{0.77 $\pm$ 0.16} & 0.83 $\pm$ 0.21 & 0.67 $\pm$ 0.20 \\
Motion Blur & \textbf{0.96 $\pm$ 0.08} & \textbf{0.85 $\pm$ 0.11} & 0.86 $\pm$ 0.20 & 0.69 $\pm$ 0.21 \\
Rain & 0.24 $\pm$ 0.14 & 0.09 $\pm$ 0.05 & \textbf{0.89 $\pm$ 0.16} & \textbf{0.73 $\pm$ 0.18} \\
Speckle Noise & 0.89 $\pm$ 0.12 & 0.75 $\pm$ 0.16 & \textbf{0.94 $\pm$ 0.09} & \textbf{0.81 $\pm$ 0.13} \\
\bottomrule
\end{tabular}
}
\end{table}

\subsection{Testing FEP-Nav on a real-world robot}
\textbf{Continuous-action Navigation Model:} During data collection, reflective markers are attached to the drone, and motion capture cameras are used to track its position and orientation. Instead of performing actual flights, the drone is manually moved by humans from the starting point to the target point to simulate its movement.
During testing, the drone's velocity is used to estimate its odometry. We collected around 300 episodes of data in a single room, which amounts to approximately 3 hours of data. The navigation model we use has the same architecture as the one implemented in the Habitat simulation. The \emph{Visual Encoder (VE)} is initialised with the Habitat navigation model weights, but the policy is initialised from scratch due to differences in the action space. The output action from the model is $a_{t} = (x_{t+1} - x_{t}, y_{t+1} - y_{t}, z_{t+1} - z_{t}, \theta_{t+1} - \theta_{t})$ \cite{shah2023gnm}, where $x, y, z$ are in the drone's coordinate system, and $\theta$ represents the yaw angle. The navigation model is fine-tuned using behavioural cloning with $L1$ loss between the predicted action and the ground-truth action. We refer to the model fine-tuned from the Pretrained-Nav weights as the Vanilla Policy. During testing, we omit the $z$ action because we do not train it to fly at different altitudes.

\textbf{Evaluation Task:} The drone is placed in front of the suitcase at a distance of approximately 0.4 to 1 meter. The drone must navigate to the backside of the suitcase without colliding with it to successfully complete an episode. We use the Success Rate (SR) as the performance metric.

\begin{table}[t]
\vspace{3mm}
\centering
\caption{\textbf{Ablation Study:} Applying AN to different parts of the model}
\label{tab:ablation}
\vspace{-3mm}
\resizebox{0.8\columnwidth}{!}
{
\begin{tabular}{lcccc}
\toprule
\multirow{2}{*}{Configuration} & \multicolumn{2}{c}{Average} & \multicolumn{2}{c}{Minimum} \\
\cmidrule(lr){2-3} \cmidrule(lr){4-5}
 & SR & SPL & SR & SPL \\
\midrule
TD-Only & 0.75 $\pm$ 0.20 & 0.55 $\pm$ 0.23 & 0.30 $\pm$ 0.20 & 0.11 $\pm$0.23 \\
AN-Only & 0.89 $\pm$ 0.07 & 0.73 $\pm$ 0.11 & 0.76 $\pm$ 0.07 & 0.51 $\pm$ 0.11 \\
TD+AN: Block 1 & 0.84 $\pm$ 0.22 & 0.70 $\pm$ 0.24 & 0.27 $\pm$ 0.22 & 0.10 $\pm$ 0.24 \\
TD+AN: Block 2 & 0.84 $\pm$ 0.20 & 0.69 $\pm$ 0.22 & 0.34 $\pm$ 0.20 & 0.14 $\pm$ 0.22 \\
TD+AN: Block 3 & 0.84 $\pm$ 0.18 & 0.68 $\pm$ 0.20 & 0.40 $\pm$ 0.18 & 0.18 $\pm$ 0.20 \\
TD+AN: All (FEP-Nav) & \textbf{0.92 $\pm$ 0.05} & \textbf{0.79 $\pm$ 0.07} & \textbf{0.83 $\pm$ 0.05} & \textbf{0.67 $\pm$ 0.07} \\

\bottomrule
\end{tabular}
}
\vspace{-3mm}
\end{table}

\begin{figure}[t]
\vspace{2mm}
\centering
\includegraphics[scale=0.32]{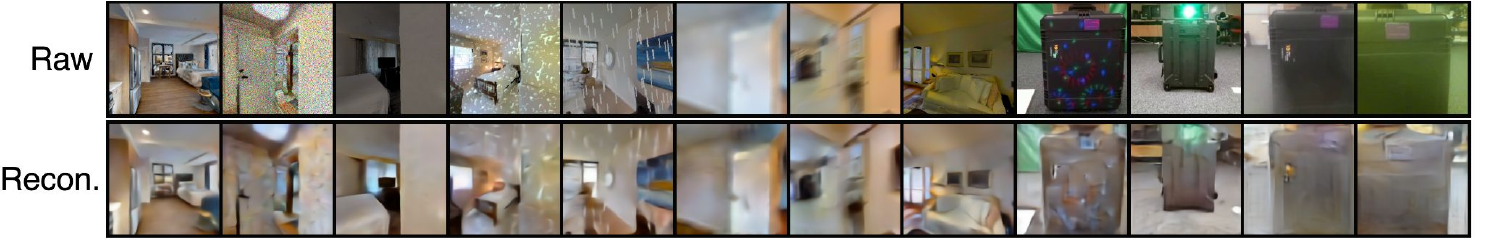}
\vspace{-2mm}
\caption{\textbf{Top-down Decoder (TD) with Adaptive Normalisation in FEP-Nav can reconstruct images affected by unseen corruptions:} The top-row images are inputs to Visual Encoder (VE). The bottom-row images are reconstructions from the same VE's inputs.}
\vspace{-4mm}
\label{fig:recon}
\end{figure}

\subsection{Results}
\label{subsec:results}
\textbf{Adaptive Normalisation denoises images:} Fig. \ref{fig:recon} shows that the reconstruction from \emph{TD} is brighter, sharper, and clearer than the original inputs which are dimmed, blurred, and noisy, respectively. Additionally, the reconstruction contains fewer water and rain droplets compared to the raw images. Furthermore, the colour in the reconstruction returns to normal when the input image is tinted yellow.

\textbf{Adaptive Normalisation reduces empirical KL:}
To test whether \emph{AN} reduces the Bayesian surprise term in Eq.~\eqref{eq:elbo_decomposition}, we estimate the latent KL divergence between corrupted and clean feature distributions. We collect clean rollout frames, apply each visual corruption offline to the same frames, and extract VE features before the recurrent policy with and without \emph{AN}. After flattening feature tensors, standardising pooled clean/corrupted samples, and applying PCA, we estimate $D_{KL}(Q_{\text{corr}}(Z)\|P_{\text{clean}}(Z))$ using a non-parametric $k$-nearest-neighbour estimator. KL is computed per scene and reported as mean $\pm$ scene standard deviation; lower KL means the corrupted representation is closer to the clean distribution.

While Proposition~\ref{thm:reduce_kl} assumes a standard Gaussian prior for analytical tractability, the empirical analysis uses the latent-feature distribution induced by clean observations as the corresponding data-driven prior. As shown in Table~\ref{tab:kl_summary_pre_lstm}, \emph{AN} yields lower mean empirical KL for seven out of eight visual corruptions. The largest reductions occur for \textit{Rain}, \textit{Light Out}, \textit{Lighting}, and \textit{Speckle Noise}, indicating that \emph{AN} substantially aligns corrupted latent features with clean features. For \textit{Defocus Blur}, the reduction is small relative to the observed variability, while for \textit{Motion Blur}, the KL values are nearly identical. Overall, these results support the Bayesian-surprise interpretation of \emph{AN}.

\textbf{FEP-Nav outperforms other methods on navigation under visual corruptions:} 
As shown in \autoref{table:performance}, FEP-Nav Instance shows improvement over all baseline methods, achieving at least $83\%$ SR and $67\%$ SPL across all individual corruptions, and achieving at least $89\%$ SR on three of the most severe corruptions (\textit{Speckle Noise}, \textit{Lighting}, and \textit{Rain}). Additionally, FEP-Nav Instance achieves the highest overall average SR ($92\%$) and SPL ($79\%$) across all types of visual observations\footnote{As shown in \autoref{table:performance}, \textit{Speckle Noise}, \textit{Lighting}, \textit{Spatter}, \textit{Rain}, and \textit{Defocus Blur} can be classified as severe corruptions due to the substantial performance degradation observed.\label{severe}}. FEP-Nav Instance ranks first or second across all non-blur corruption types. Crucially, its overall average and worst-case minimum performance (lower bound) are the best among all evaluation methods. A remaining limitation is performance under strong blur corruptions, where high-frequency visual details are lost and cannot be fully recovered by feature normalisation or the current reconstruction module. This suggests that stronger restoration-oriented decoders may further improve the quality of surrogate observations and, consequently, test-time adaptation under blur.

\begin{figure}[t]
\vspace{+0.2cm}
\centering
\includegraphics[scale=0.25]{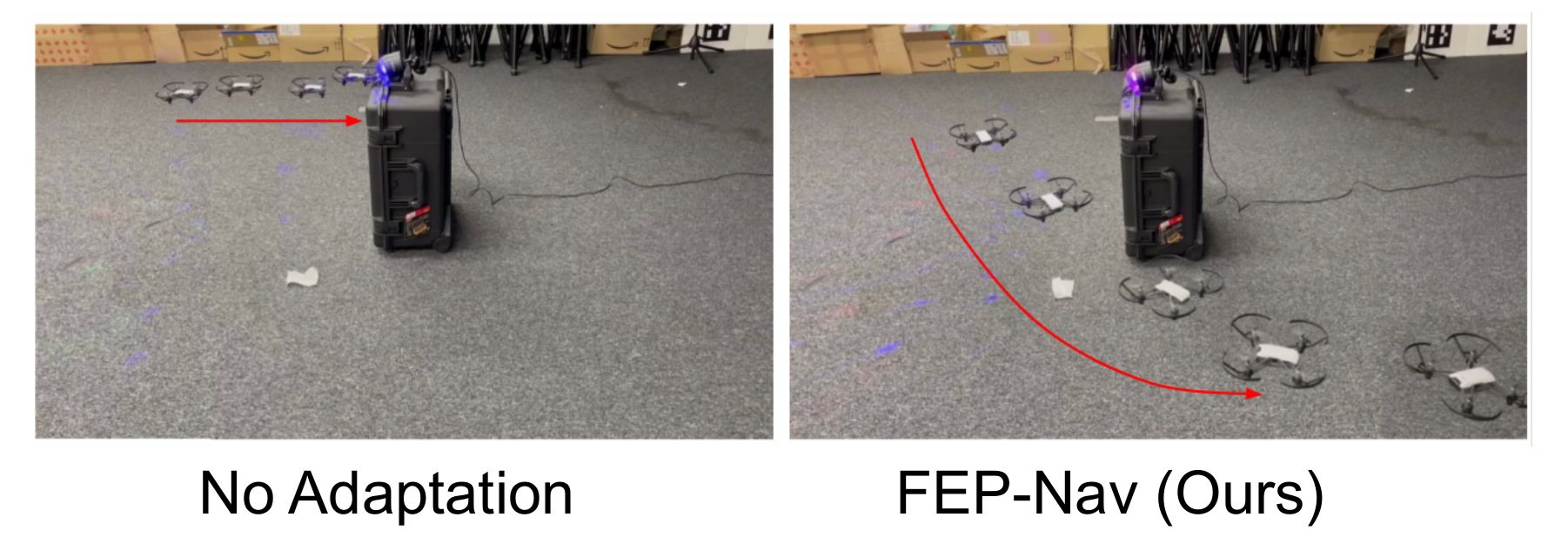}
\vspace{-0.4cm}
\caption{\textbf{FEP-Nav helps the robot avoid the obstacle when the drone's camera is perturbed by lights:} (Left) The drone with the non-adaptive navigation policy crashes into the suitcase when the disco light is applied to the drone camera. (Right) The FEP-Nav Instance allows the drone to adapt to the visual corruption and successfully avoid the suitcase. Red curve indicates the drone trajectory.}
\label{fig:drone_path}
\vspace{-0.4cm}
\end{figure}

\textbf{FEP-Nav helps the robot avoid the obstacle:} We apply two TTA methods that work well in the simulation, DUA and FEP-Nav Instance, to the Vanilla Policy, which is fine-tuned from the Pretrained-Nav model using behavioural cloning. \emph{TD} is trained on both 112k Gibson images and additional real-world images, which are used to train the drone navigation model. FEP-Nav Instance outperforms DUA across all types of visual corruption, as shown in Table \ref{tab:drone_results}. Interestingly, DUA barely improves the success rate over the Vanilla Policy when facing dirt on the camera, while FEP-Nav maintains a high success rate of $80\%$. Therefore, FEP-Nav Instance is the best among evaluated methods for handling real-world corruption types. \autoref{fig:drone_path} shows an example of the navigation trajectory with and without adaptation.

\textbf{Comparison with Image Restoration:} We evaluate FEP-Nav against MPRNet \cite{zamir2021multi}, a strong model for image restoration. While MPRNet is explicitly trained on clean-corrupt image pairs to map degraded inputs back to a clean ground truth, its efficacy remains constrained by the specific noise distributions present in its training data. As shown in Table~\ref{tab:restoration_comparison}, MPRNet underperforms our method on \emph{Rain} and \emph{Speckle Noise} corruptions. We attribute this to a distribution shift: although the simulated corruptions may visually resemble the model's training set, a neural network may perceive the underlying statistical noise differently, leading to suboptimal restoration. In contrast, FEP-Nav does not require paired clean-corrupt data. By training solely on uncorrupted frames, our method treats all sensory shifts as out-of-distribution challenges to be resolved at test time. This independence from corruption-specific priors allows FEP-Nav to maintain functional stability across a broader spectrum of unforeseen noise types where specialised restoration models fail to generalise. 

\textbf{FEP-Nav allows real-time adaptation:} FEP-Nav requires an additional inference stage, which increases computational overhead. The additional encoding and decoding require $0.045$s per frame and around $150$ MB of graphics memory on the NVIDIA Jetson AGX Orin 32GB. Due to Tello's limited payload capacity, computation was performed at a ground station.

\textbf{FEP-Nav Ablations:}
We evaluate the contributions of the TD and AN, and further study which normalisation layers benefit most from adaptation.
\textit{TD-Only} uses the decoder without updating BatchNorm statistics, while \textit{AN-Only} adapts BatchNorm statistics during inference without using the decoder.
\textit{TD+AN: Block 1}, \textit{TD+AN: Block 2}, and \textit{TD+AN: Block 3} use both TD and AN, but update BatchNorm statistics only in the corresponding VE block while keeping the other blocks frozen.
\textit{TD+AN: All (FEP-Nav)} updates BatchNorm statistics across all VE blocks and corresponds to the full proposed method. As shown in \autoref{tab:ablation}, \textit{AN-Only} substantially outperforms \textit{TD-Only}, indicating that AN is the main contributor to robustness under visual corruption. Nevertheless, \textit{TD+AN: All (FEP-Nav)} achieves the best average and minimum SR/SPL, showing that TD provides a complementary benefit.

\section{Conclusion}

We present FEP-Nav, an \emph{FEP-inspired} real-time adaptation framework for robust visual navigation. Motivated by the decomposition of Variational Free Energy (VFE) into prediction error and Bayesian surprise, FEP-Nav combines a Top-down Decoder and Adaptive Normalisation to mitigate sensory non-stationarity without requiring online gradient updates.
Evaluations across simulated and real-world corruptions demonstrate that FEP-Nav substantially recovers navigation performance, outperforming strong adaptive methods. The system's computational efficiency ensures its viability for real-time deployment on embedded robotic platforms, such as NVIDIA Jetson AGX Orin. Our results suggest that grounding robotic perception in these generative formalisms offers a principled pathway towards robust autonomous navigation under real-world sensory degradation.

\appendix
Proof of Proposition~\ref{thm:reduce_kl}

Let $J(\mu, \sigma) = D_{KL}(\mathcal{N}(\mu, \sigma^2) \| \mathcal{N}(0, 1))$ denote the divergence from the prior. We analyse the behaviour of $J$ under two conditions: unnormalised (corrupted) and normalised (adapted). Visual corruption induces a domain shift, causing the true moments of $z$ to deviate significantly from the prior. The divergence is:
\begin{equation}
\text{RHS} = J(\mu_{corr}, \sigma_{corr}) = \frac{1}{2}\left(\mu_{corr}^2 + \sigma_{corr}^2 - 1 - \ln(\sigma_{corr}^2)\right)
\end{equation}

We assume the corruption is significant ($|\mu_{corr}| \gg 0$ and $|\sigma_{corr}^2 - 1| \gg 0$), resulting in a large divergence value ($\text{RHS} \gg 0$). We normalise $z$ using empirical statistics $\hat{\mu}$ and $\hat{\sigma}$. Since the spatial resolution $N = H \times W$ is large, the statistical estimation errors $\epsilon$ are negligible relative to unity ($|\epsilon| \ll 1$). We approximate the normalised moments as $\tilde{\mu} \approx -\epsilon_\mu$ and $\tilde{\sigma} \approx 1 - \epsilon_\sigma$. The divergence becomes:
\begin{align}
\text{LHS} &= \frac{1}{2} \left[ \tilde{\mu}^2 + \tilde{\sigma}^2 - 1 - \ln(\tilde{\sigma}^2) \right] \\
&\approx \frac{1}{2} \left[ (-\epsilon_\mu)^2 + (1-\epsilon_\sigma)^2 - 1 - 2\ln(1-\epsilon_\sigma) \right]
\end{align}
Using the Taylor expansion $\ln(1-x) \approx -x - x^2/2$, which is valid given $|x| \ll 1$:
\begin{align}
\text{LHS} &\approx \frac{1}{2} \left[ \epsilon_\mu^2 + 2\epsilon_\sigma^2 \right]
\end{align}

The LHS is of second order in the tiny error term ($\mathcal{O}(\epsilon^2)$), effectively approaching zero. In contrast, the RHS is of second order in the significant corruption shift ($\mathcal{O}(\Delta^2)$).
Since $|\epsilon| \ll \Delta$, it follows that $\text{LHS} \ll \text{RHS}$. Thus, Adaptive Normalisation reduces the Bayesian surprise.
\vspace{-3mm}
\section*{Acknowledgment}
The authors thank the anonymous reviewers for their constructive feedback. They also acknowledge Research IT and the Computational Shared Facility at The University of Manchester.
\vspace{-2mm}

\bibliographystyle{IEEEtran}
\bibliography{ref}

@inproceedings{mezghan2022memory,
  title={Memory-augmented reinforcement learning for image-goal navigation},
  author={Mezghan, Lina and Sukhbaatar, Sainbayar and Lavril, Thibaut and Maksymets, Oleksandr and Batra, Dhruv and Bojanowski, Piotr and Alahari, Karteek},
  booktitle={2022 IEEE/RSJ International Conference on Intelligent Robots and Systems (IROS)},
  pages={3316--3323},
  year={2022},
  organization={IEEE}
}

@inproceedings{habitat19iccv,
  title     =     {Habitat: {A} {P}latform for {E}mbodied {AI} {R}esearch},
  author    =     {{Manolis Savva*} and {Abhishek Kadian*} and {Oleksandr Maksymets*} and Yili Zhao and Erik Wijmans and Bhavana Jain and Julian Straub and Jia Liu and Vladlen Koltun and Jitendra Malik and Devi Parikh and Dhruv Batra},
  booktitle =     {Proceedings of the IEEE/CVF International Conference on Computer Vision (ICCV)},
  year      =     {2019}
}

@inproceedings{xiazamirhe2018gibsonenv,
  title={Gibson {Env}: real-world perception for embodied agents},
  author={Xia, Fei and R. Zamir, Amir and He, Zhi-Yang and Sax, Alexander and Malik, Jitendra and Savarese, Silvio},
  booktitle={Computer Vision and Pattern Recognition (CVPR), 2018 IEEE Conference on},
  year={2018},
  organization={IEEE}
}

@inproceedings{wijmans2019ddppo,
  title={DD-PPO: Learning Near-Perfect PointGoal Navigators from 2.5 Billion Frames},
  author={Wijmans, Erik and Kadian, Abhishek and Morcos, Ari and Lee, Stefan and Essa, Irfan and Parikh, Devi and Savva, Manolis and Batra, Dhruv},
  booktitle={International Conference on Learning Representations},
  year={2019}
}

@inproceedings{wu2018groupnrom,
  title={Group normalization},
  author={Wu, Yuxin and He, Kaiming},
  booktitle={Proceedings of the European conference on computer vision (ECCV)},
  pages={3--19},
  year={2018}
}

@inproceedings{rajivc2022robustness,
  title={Robustness of Embodied Point Navigation Agents},
  author={Raji{\v{c}}, Frano},
  booktitle={European Conference on Computer Vision},
  pages={193--204},
  year={2022},
  organization={Springer}
}

@inproceedings{mirza2022dua,
  title={The norm must go on: Dynamic unsupervised domain adaptation by normalization},
  author={Mirza, M Jehanzeb and Micorek, Jakub and Possegger, Horst and Bischof, Horst},
  booktitle={Proceedings of the IEEE/CVF Conference on Computer Vision and Pattern Recognition},
  pages={14765--14775},
  year={2022}
}

@inproceedings{wang2020tent,
  title={Tent: Fully Test-Time Adaptation by Entropy Minimization},
  author={Wang, Dequan and Shelhamer, Evan and Liu, Shaoteng and Olshausen, Bruno and Darrell, Trevor},
  booktitle={International Conference on Learning Representations},
  year={2020}
}

@inproceedings{liang2020shot,
  title={Do we really need to access the source data? source hypothesis transfer for unsupervised domain adaptation},
  author={Liang, Jian and Hu, Dapeng and Feng, Jiashi},
  booktitle={International conference on machine learning},
  pages={6028--6039},
  year={2020},
  organization={PMLR}
}

@article{ulyanov2016instance,
  title={Instance normalization: The missing ingredient for fast stylization},
  author={Ulyanov, Dmitry and Vedaldi, Andrea and Lempitsky, Victor},
  journal={arXiv preprint arXiv:1607.08022},
  year={2016}
}

@inproceedings{hu2018squeeze,
  title={Squeeze-and-excitation networks},
  author={Hu, Jie and Shen, Li and Sun, Gang},
  booktitle={Proceedings of the IEEE conference on computer vision and pattern recognition},
  pages={7132--7141},
  year={2018}
}

@inproceedings{lin2023vitta,
  title={Video Test-Time Adaptation for Action Recognition},
  author={Lin, Wei and Mirza, Muhammad Jehanzeb and Kozinski, Mateusz and Possegger, Horst and Kuehne, Hilde and Bischof, Horst},
  booktitle={Proceedings of the IEEE/CVF Conference on Computer Vision and Pattern Recognition},
  pages={22952--22961},
  year={2023}
}

@inproceedings{wang2023diga,
  title={Dynamically Instance-Guided Adaptation: A Backward-Free Approach for Test-Time Domain Adaptive Semantic Segmentation},
  author={Wang, Wei and Zhong, Zhun and Wang, Weijie and Chen, Xi and Ling, Charles and Wang, Boyu and Sebe, Nicu},
  booktitle={Proceedings of the IEEE/CVF Conference on Computer Vision and Pattern Recognition},
  pages={24090--24099},
  year={2023}
}

@inproceedings{child2020very,
  title={Very Deep VAEs Generalize Autoregressive Models and Can Outperform Them on Images},
  author={Child, Rewon},
  booktitle={International Conference on Learning Representations},
  year={2020}
}

@inproceedings{zhu2017rl_nav,
  title={Target-driven visual navigation in indoor scenes using deep reinforcement learning},
  author={Zhu, Yuke and Mottaghi, Roozbeh and Kolve, Eric and Lim, Joseph J and Gupta, Abhinav and Fei-Fei, Li and Farhadi, Ali},
  booktitle={2017 IEEE international conference on robotics and automation (ICRA)},
  pages={3357--3364},
  year={2017},
  organization={IEEE}
}

@article{anderson2018evaluation,
  title={On evaluation of embodied navigation agents},
  author={Anderson, Peter and Chang, Angel and Chaplot, Devendra Singh and Dosovitskiy, Alexey and Gupta, Saurabh and Koltun, Vladlen and Kosecka, Jana and Malik, Jitendra and Mottaghi, Roozbeh and Savva, Manolis and others},
  journal={arXiv preprint arXiv:1807.06757},
  year={2018}
}

@article{pandey2022diffusevae,
  title={DiffuseVAE: Efficient, Controllable and High-Fidelity Generation from Low-Dimensional Latents},
  author={Pandey, Kushagra and Mukherjee, Avideep and Rai, Piyush and Kumar, Abhishek},
  journal={Transactions on Machine Learning Research},
  year={2022}
}

@article{kreiman2020beyond,
  title={Beyond the feedforward sweep: feedback computations in the visual cortex},
  author={Kreiman, Gabriel and Serre, Thomas},
  journal={Annals of the New York Academy of Sciences},
  volume={1464},
  number={1},
  pages={222--241},
  year={2020},
  publisher={Wiley Online Library}
}

@inproceedings{shah2023gnm,
  title={Gnm: A general navigation model to drive any robot},
  author={Shah, Dhruv and Sridhar, Ajay and Bhorkar, Arjun and Hirose, Noriaki and Levine, Sergey},
  booktitle={2023 IEEE International Conference on Robotics and Automation (ICRA)},
  pages={7226--7233},
  year={2023},
  organization={IEEE}
}

@article{li2016revisiting,
  title={Revisiting batch normalization for practical domain adaptation},
  author={Li, Yanghao and Wang, Naiyan and Shi, Jianping and Liu, Jiaying and Hou, Xiaodi},
  journal={arXiv preprint arXiv:1603.04779},
  year={2016}
}

@inproceedings{kang2024membn,
  title={MemBN: Robust Test-Time Adaptation via Batch Norm with Statistics Memory},
  author={Kang, Juwon and Kim, Nayeong and Ok, Jungseul and Kwak, Suha},
  booktitle={European Conference on Computer Vision},
  pages={467--483},
  year={2024},
  organization={Springer}
}

@inproceedings{kim2025test,
  title={Test-Time Adaptation for Online Vision-Language Navigation with Feedback-based Reinforcement Learning},
  author={Kim, Sungjune and Oh, Gyeongrok and Ko, Heeju and Ji, Daehyun and Lee, Dongwook and Lee, Byung-Jun and Jang, Sujin and Kim, Sangpil},
  booktitle={International Conference on Machine Learning},
year={2025}
}

@inproceedings{ko2025active,
  title={Active Test-time Vision-Language Navigation},
  author={Ko, Heeju and Kim, Sungjune and Oh, Gyeongrok and Yoon, Jeongyoon and Lee, Honglak and Jang, Sujin and Kim, Seungryong and Kim, Sangpil},
  booktitle={Advances in Neural Information Processing Systems},
  year={2025},
}

@article{rao1999predictive,
  title={Predictive coding in the visual cortex: a functional interpretation of some extra-classical receptive-field effects},
  author={Rao, Rajesh PN and Ballard, Dana H},
  journal={Nature neuroscience},
  volume={2},
  number={1},
  pages={79--87},
  year={1999},
  publisher={Nature Publishing Group}
}

@article{friston2010free,
  title={The free-energy principle: a unified brain theory?},
  author={Friston, Karl},
  journal={Nature reviews neuroscience},
  volume={11},
  number={2},
  pages={127--138},
  year={2010},
  publisher={Nature publishing group}
}

@inproceedings{meo2021multimodal,
  title={Multimodal vae active inference controller},
  author={Meo, Cristian and Lanillos, Pablo},
  booktitle={2021 IEEE/RSJ International Conference on Intelligent Robots and Systems (IROS)},
  pages={2693--2699},
  year={2021},
  organization={IEEE}
}

@inproceedings{bos2022free,
  title={Free energy principle for state and input estimation of a quadcopter flying in wind},
  author={Bos, Fred and Meera, Ajith Anil and Benders, Dennis and Wisse, Martijn},
  booktitle={2022 International Conference on Robotics and Automation (ICRA)},
  pages={5389--5395},
  year={2022},
  organization={IEEE}
}

@inproceedings{baioumy2022unbiased,
  title={Unbiased active inference for classical control},
  author={Baioumy, Mohamed and Pezzato, Corrado and Ferrari, Riccardo and Hawes, Nick},
  booktitle={2022 IEEE/RSJ International Conference on Intelligent Robots and Systems (IROS)},
  pages={12787--12794},
  year={2022},
  organization={IEEE}
}

@inproceedings{lanillos2018adaptive,
  title={Adaptive robot body learning and estimation through predictive coding},
  author={Lanillos, Pablo and Cheng, Gordon},
  booktitle={2018 IEEE/RSJ International Conference on Intelligent Robots and Systems (IROS)},
  pages={4083--4090},
  year={2018},
  organization={IEEE}
}

@inproceedings{baioumy2021active,
  title={Active inference for integrated state-estimation, control, and learning},
  author={Baioumy, Mohamed and Duckworth, Paul and Lacerda, Bruno and Hawes, Nick},
  booktitle={2021 IEEE International Conference on Robotics and Automation (ICRA)},
  pages={4665--4671},
  year={2021},
  organization={IEEE}
}

@article{meo2022adaptation,
  title={Adaptation through prediction: Multisensory active inference torque control},
  author={Meo, Cristian and Franzese, Giovanni and Pezzato, Corrado and Spahn, Max and Lanillos, Pablo},
  journal={IEEE Transactions on Cognitive and Developmental Systems},
  volume={15},
  number={1},
  pages={32--41},
  year={2022},
  publisher={IEEE}
}

@article{lanillos2021active,
  title={Active inference in robotics and artificial agents: Survey and challenges},
  author={Lanillos, Pablo and Meo, Cristian and Pezzato, Corrado and Meera, Ajith Anil and Baioumy, Mohamed and Ohata, Wataru and Tschantz, Alexander and Millidge, Beren and Wisse, Martijn and Buckley, Christopher L and others},
  journal={arXiv preprint arXiv:2112.01871},
  year={2021}
}

@book{parr2022active,
  title={Active inference: the free energy principle in mind, brain, and behavior},
  author={Parr, Thomas and Pezzulo, Giovanni and Friston, Karl J},
  year={2022},
  publisher={MIT Press}
}

@article{da2022active,
  title={How active inference could help revolutionise robotics},
  author={Da Costa, Lancelot and Lanillos, Pablo and Sajid, Noor and Friston, Karl and Khan, Shujhat},
  journal={Entropy},
  volume={24},
  number={3},
  pages={361},
  year={2022},
  publisher={MDPI}
}

@article{mazzaglia2021contrastive,
  title={Contrastive active inference},
  author={Mazzaglia, Pietro and Verbelen, Tim and Dhoedt, Bart},
  journal={Advances in neural information processing systems},
  volume={34},
  pages={13870--13882},
  year={2021}
}

@article{fountas2020deep,
  title={Deep active inference agents using Monte-Carlo methods},
  author={Fountas, Zafeirios and Sajid, Noor and Mediano, Pedro and Friston, Karl},
  journal={Advances in neural information processing systems},
  volume={33},
  pages={11662--11675},
  year={2020}
}

@inproceedings{van2020deep,
  title={Deep active inference for partially observable MDPs},
  author={van der Himst, Otto and Lanillos, Pablo},
  booktitle={International Workshop on Active Inference},
  pages={61--71},
  year={2020},
  organization={Springer}
}

@article{ueltzhoffer2018deep,
  title={Deep active inference},
  author={Ueltzh{\"o}ffer, Kai},
  journal={Biological cybernetics},
  volume={112},
  number={6},
  pages={547--573},
  year={2018},
  publisher={Springer}
}

@inproceedings{lotter2017deep,
  title={Deep Predictive Coding Networks for Video Prediction and Unsupervised Learning},
  author={Lotter, William and Kreiman, Gabriel and Cox, David},
  booktitle={International Conference on Learning Representations},
  year={2017}
}

@article{vijayaraghavan2025development,
  title={Development of compositionality through interactive learning of language and action of robots},
  author={Vijayaraghavan, Prasanna and Quei{\ss}er, Jeffrey Frederic and Flores, Sergio Verduzco and Tani, Jun},
  journal={Science Robotics},
  volume={10},
  number={98},
  pages={eadp0751},
  year={2025},
  publisher={American Association for the Advancement of Science}
}

@article{fujii2025real,
  title={Real-World Robot Control by Deep Active Inference With a Temporally Hierarchical World Model},
  author={Fujii, Kentaro and Murata, Shingo},
  journal={IEEE Robotics and Automation Letters},
  volume={11},
  number={1},
  pages={890--897},
  year={2025},
  publisher={IEEE}
}

@article{wirkuttis2021leading,
  title={Leading or following? Dyadic robot imitative interaction using the active inference framework},
  author={Wirkuttis, Nadine and Tani, Jun},
  journal={IEEE Robotics and Automation Letters},
  volume={6},
  number={3},
  pages={6024--6031},
  year={2021},
  publisher={IEEE}
}

@inproceedings{meera2023adaptive,
  title={Adaptive noise covariance estimation under colored noise using dynamic expectation maximization},
  author={Meera, Ajith Anil and Lanillos, Pablo},
  booktitle={2023 62nd IEEE Conference on Decision and Control (CDC)},
  pages={165--171},
  year={2023},
  organization={IEEE}
}

@inproceedings{zamir2021multi,
  title={Multi-stage progressive image restoration},
  author={Zamir, Syed Waqas and Arora, Aditya and Khan, Salman and Hayat, Munawar and Khan, Fahad Shahbaz and Yang, Ming-Hsuan and Shao, Ling},
  booktitle={Proceedings of the IEEE/CVF conference on computer vision and pattern recognition},
  pages={14821--14831},
  year={2021}
}

@article{kim2026buffer,
  title={Buffer layers for test-time adaptation},
  author={Kim, Hyeongyu and Han, Geonhui and Hwang, Dosik},
  journal={Advances in Neural Information Processing Systems},
  volume={38},
  pages={141127--141149},
  year={2026}
}
\end{document}